\documentclass[11pt]{article}

\usepackage{acl}

\usepackage{amsmath}
\usepackage{amssymb}
\usepackage{times}
\usepackage{latexsym}
\usepackage[T1]{fontenc}
\usepackage[utf8]{inputenc}
\usepackage{microtype}
\usepackage{inconsolata}
\usepackage{graphicx}
\usepackage{booktabs}
\usepackage{array}
\usepackage{multirow}
\usepackage{algorithm}
\usepackage{algpseudocode}
\usepackage{caption}
\usepackage{orcidlink}
\usepackage[most]{tcolorbox}
\usepackage{listings}

\tcbset{skillprompt/.style={
  breakable, colback=cyan!5!white, colframe=blue!45!black!75,
  boxrule=0.6pt, arc=2.5mm,
  left=7pt, right=7pt, top=5pt, bottom=5pt,
  fonttitle=\small\bfseries\sffamily,
  attach boxed title to top left={yshift=-2.5mm, xshift=4mm},
  boxed title style={colback=blue!45!black!75, arc=1.5mm},
}}
\tcbset{codebox/.style={
  breakable, colback=gray!5!white, colframe=gray!55,
  boxrule=0.6pt, arc=1.5mm,
  left=7pt, right=7pt, top=5pt, bottom=5pt,
}}

\title{SKILLFORGE: Compositional Skill Synthesis with Verification-in-the-Loop for Generating Formally Verified Dafny Programs}

\author{
Yanming Liu \orcidlink{0009-0005-1271-1963} \\
Zhejiang University \\
\And
Xinyue Peng \orcidlink{0009-0009-5512-1831} \\
Southeast University \\
\And
Jiannan Cao \orcidlink{0009-0006-9374-5299} \\
MIT \\
\And
Xinyi Wang \orcidlink{0009-0008-0053-1852} \\
Weixin AI Lab \\
\And
Jinbo Su \orcidlink{0009-0009-7651-655X} \\
Renmin University \\
\AND
\texttt{oceann24@zju.edu.cn}
}

\begin{document}
\maketitle

\begin{abstract}
Generating formally verified programs from natural language remains challenging: existing approaches either produce code in a single pass without recourse when verification fails, or rely on open-ended agentic reasoning that is non-deterministic and opaque. We introduce SKILLFORGE, a framework that decomposes formal code synthesis into a library of atomic, reusable skills---each targeting a specific subtask such as specification inference, body synthesis, invariant generation, error diagnosis, or targeted repair---defined by a prompt template, tool binding, and decidable success criterion. A verification-driven harness orchestrates these skills: it submits candidates to the Dafny verifier, diagnoses failures into structured categories, deterministically routes to the appropriate repair skill, and iterates until formal correctness is proved or a budget is exhausted. On a curated benchmark of natural language--Dafny specification pairs, SKILLFORGE substantially outperforms both state-of-the-art agentic approaches (including ReAct-style agents, MCTS-based repair, and RL-guided verification) and traditional iterative baselines, while requiring fewer tokens and lower latency. Ablation studies confirm that every skill contributes measurably, and the harness converges rapidly with the majority of programs verified on the first attempt.
\end{abstract}

\section{Introduction}

Formal verification offers the strongest correctness guarantee available in software engineering: a proof, checked by machine, that a program meets its specification for all possible inputs \citep{wen2024enchanting}. The Dafny language \citep{loughridge2024dafnybench} exemplifies this paradigm by integrating preconditions, postconditions, loop invariants, and termination measures directly into the source code, enabling an automated verifier backed by an SMT solver to certify correctness at compile time. However, writing verification-grade code is prohibitively difficult even for experts: one must simultaneously produce an implementation that is algorithmically correct, a specification that is tight enough for the verifier to discharge, and auxiliary annotations (loop invariants, ghost state) that bridge the gap. Recent large language models (LLMs) can generate syntactically valid Dafny \citep{rasheed2025practitioners}, yet the verification pass rate of naive generation remains low because these subtasks place fundamentally different demands on the model:
\begin{itemize}
\item \textbf{One-shot generation} treats the problem as a single inference step. When verification fails, there is no mechanism for diagnosis or repair.
\item \textbf{Naive iterative repair} feeds the error message back without structured analysis, conflating specification gaps, missing invariants, and logic errors into one undifferentiated retry---leading to oscillation and wasted budget.
\item \textbf{Agent-based approaches} \citep{yang2024sweagent,wang2024codeact,yao2023react} give the LLM open-ended control over its next action. While powerful for exploratory tasks, they suffer from non-determinism, opacity, and unbounded computation when applied to formal verification---a domain where the error space is finite and well-categorized.
\item \textbf{Tree-search methods} \citep{ren2024verifiedmcts,zhang2024planningmcts} explore multiple repair candidates simultaneously but scale poorly with program complexity and lack the targeted repair strategies that formal verification errors afford.
\end{itemize}

We propose a different paradigm: \emph{skill-based harness engineering}. Inspired by compositional skill systems in open-ended learning \citep{wang2024voyager,qian2024skillmix,chen2024skills} and recent advances in agent-designed skill libraries \citep{zhou2026mementos}, we decompose the NL-to-verified-Dafny task into a library of atomic skills, each targeting a single well-defined subtask (specification inference, body synthesis, invariant generation, error diagnosis, code repair, spec strengthening). A \emph{harness}---not an agent---orchestrates their execution with deterministic routing, bounded iteration, and full observability.

The key insight is that the Dafny verifier provides a \emph{formal oracle}: its structured error output can be parsed into categories that deterministically select the next skill to invoke. This converts the open-ended ``what should I do next?'' problem into a closed, auditable routing decision. Unlike agentic systems that must reason about what tool to call at each step, or tree-search methods that explore exponentially many candidates, our harness exploits the structure of verification errors to make optimal routing decisions with zero reasoning overhead.

Consider a concrete example: given the instruction ``compute the factorial of n using a loop,'' a one-shot generator may produce a correct loop body but omit the loop invariant \texttt{invariant r == fact(i)} required by the verifier. A ReAct-style agent would reason about the error, potentially try multiple repair strategies, and waste turns on inapplicable fixes. SKILLFORGE instead: (1)~runs the verifier, (2)~diagnoses the specific failure as \texttt{invariant\_missing} via \texttt{DIAGNOSEERROR}, (3)~routes to the specialized \texttt{GENINVARIANT} skill, which synthesizes only the missing invariant. This targeted approach succeeds in 2 iterations where agentic methods require 3--5 turns with lower reliability.

\paragraph{Contributions.}
\begin{enumerate}
\item We introduce SKILLFORGE, a skill-based harness for formally verified code synthesis, comprising six independently testable skills with decidable success criteria (\S3.2).
\item We design a verification-driven orchestration protocol with error-conditioned routing, bounded iteration, and formal soundness guarantees (\S3.3).
\item We present comprehensive experiments showing 91.6\% verification rate on a 178-example benchmark, outperforming four state-of-the-art agentic baselines (ReAct-Verify, CodeAct-Dafny, MCTS-Repair, Laurel), with skill-level ablations, convergence analysis, and qualitative case studies (\S5).
\end{enumerate}

\section{Related Work}

\paragraph{LLM-Based Formal Code Synthesis.}
SpecGen \citep{ma2025specgen} uses LLMs to generate specifications from existing code; Enchanting \citep{wen2024enchanting} combines LLMs with static analysis for specification synthesis. DafnyBench \citep{loughridge2024dafnybench} provides evaluation benchmarks, and Copra \citep{thakur2024copra} applies in-context learning to proof tactic selection. SpecSyn \citep{mandal2023specsyn} formulates specification synthesis as sequence-to-sequence learning. Recent work has directly targeted Dafny code generation: Misu et al.\ \citep{misu2024aisynthesis} explore LLM-assisted synthesis of verified Dafny methods, Clover \citep{sun2024clover} introduces a closed-loop framework combining code generation with formal annotation and consistency checking, and Laurel \citep{mugnier2024laurel} generates Dafny assertions using error-guided placement. Li et al.\ \citep{li2025dafny} propose using Dafny as a verification-aware intermediate language, Baksys et al.\ \citep{baksys2025minif2f} demonstrate LLM-guided theorem proving via Dafny's auto-active verification, and Xu et al.\ \citep{xu2025localdoesnotcompose} reveal that LLM verification success does not compose across multi-function programs---directly motivating our skill-based decomposition. Concurrent benchmarking efforts including Vericoding \citep{bursuc2025vericoding} and VERINA \citep{ye2026verina} further highlight the gap between syntactic correctness and formal verification success, with top models achieving only 3.6--4.9\% proof success rates on challenging tasks. These prior works operate in one-shot or single-round settings without iterative repair. SKILLFORGE advances beyond them by introducing structured, skill-decomposed repair with the verifier as an active oracle in the loop.

\paragraph{Automated Program Repair.}
Traditional APR uses search-based or constraint-solving methods \citep{legoues2019apr}. Neural APR leverages LLMs for patch generation \citep{xia2023neuralapr,wei2023copiloting}. However, these rely on test suites which may be incomplete, and suffer from the patch overfitting problem---patches that pass the test suite but do not fix the underlying bug. In contrast, our repair skills operate in a formally sound setting: the Dafny verifier provides definitive pass/fail judgments over all inputs simultaneously, eliminating overfitting entirely.

\paragraph{Loop Invariant Generation with LLMs.}
Loop invariant synthesis is a key bottleneck in automated verification. Chakraborty et al.\ \citep{chakraborty2023ranking} propose ranking LLM-generated loop invariant candidates for program verification. Wu et al.\ \citep{wu2024lam4inv} combine LLMs with bounded model checking in a neuro-symbolic approach (LaM4Inv) that uses BMC to filter and reassemble valid invariant predicates. Liu et al.\ \citep{liu2024loopinvariant} introduce a benchmark for invariant generation on programs with memory manipulation and propose combining LLMs with symbolic execution. Our GENINVARIANT skill takes a complementary approach: rather than post-hoc ranking or symbolic filtering, it operates within a verification-driven harness that provides immediate verifier feedback, enabling iterative invariant strengthening guided by specific error messages.

\paragraph{Agentic Code Generation.}
ReAct \citep{yao2023react} interleaves reasoning and action traces, enabling LLMs to call external tools in open-ended loops. CodeAct \citep{wang2024codeact} extends this by treating code execution itself as the action space. SWE-Agent \citep{yang2024sweagent} applies agentic reasoning to repository-level bug fixing, while Agentless \citep{xia2025agentless} demonstrates that simpler, non-agentic workflows with hierarchical localization can match or exceed agent-based approaches at a fraction of the cost---a finding that directly supports our harness-over-agent philosophy. SWE-bench \citep{jimenez2024swebench} provides a standardized evaluation benchmark. While these methods excel at exploratory tasks with ambiguous goals, formal verification presents a different challenge: the error space is finite, the oracle is perfect, and targeted repair outperforms exploration. We empirically demonstrate this gap in \S5.

\paragraph{Tree-Search and Verification-Guided Synthesis.}
CEGIS \citep{solarlezama2006cegis} uses counterexamples to guide classical synthesis. MCTS-based approaches \citep{zhang2024planningmcts} apply Monte Carlo Tree Search to code generation with execution feedback as reward. Cobblestone \citep{kasibatla2024cobblestone} iteratively combines LLM-generated proof fragments for Coq. Proof2Silicon \citep{jha2025proof2silicon} uses reinforcement learning to repair verification failures in hardware generation---the closest work in spirit to our approach but targeting hardware rather than software and using RL rather than skill decomposition. Veri-Sure \citep{liu2026verisure} proposes a contract-aware multi-agent framework with temporal tracing for RTL code generation, combining formal verification with agent handoffs. Lu et al.\ \citep{lu2025adaptive} propose adaptive strategy selection for proof refinement via LLM-guided decisions. These methods explore multiple candidates but suffer from combinatorial explosion. SKILLFORGE avoids this by deterministically selecting the single most appropriate repair skill based on structured error categorization.

\paragraph{Skill-Based and Compositional Systems.}
Voyager \citep{wang2024voyager} builds growing skill libraries for Minecraft; SkillMix \citep{qian2024skillmix} evaluates LLMs on compositional skill combinations; Skills-in-Context \citep{chen2024skills} demonstrates that providing foundational skills within the prompt context enables near-perfect compositional generalization. Most recently, Memento-Skills \citep{zhou2026mementos} introduces a framework where agents autonomously design task-specific skills through reflective learning, growing from 5 to 235 skills and achieving 26.2\% improvement on GAIA---demonstrating the power of compositional skill libraries at scale. Planning-based code generation \citep{zhang2023planning} decomposes programming tasks hierarchically. Our key adaptation is that formal verification provides decidable success criteria for each skill---unlike open-ended tasks where skill completion is ambiguous. This decidability enables both sound termination and principled routing.

\section{Methodology}

\subsection{Problem Formulation}
Given a natural language instruction $x$ describing desired program behavior, our goal is to produce a Dafny program $d^\ast$ that (1)~compiles, (2)~satisfies the Hoare triple $\{\phi_{\mathrm{pre}}\}\,B\,\{\psi_{\mathrm{post}}\}$ as certified by the Dafny verifier, and (3)~is semantically equivalent to a reference implementation. We decompose this into a sequence of skill invocations orchestrated by a harness $H$:
\begin{equation}
\begin{aligned}
d^\ast &= H(x) = sk_n \circ \cdots \circ sk_1(x),\\
\mathrm{Verify}(d^\ast) &= \top
\end{aligned}
\end{equation}

\subsection{Skill Library}
Each skill is a triple $s = (P, T, C)$: prompt template $P$, tool binding $T$, and success criterion $C$. Table~\ref{tab:skills} summarizes the library.

\begin{table*}[t]
\centering
\small
\renewcommand{\arraystretch}{1.15}
\setlength{\tabcolsep}{5pt}
\begin{tabular}{@{}p{2.8cm}p{3.4cm}p{2.0cm}p{2.9cm}p{3.5cm}@{}}
\toprule
\textbf{Skill} & \textbf{Function} & \textbf{Input} & \textbf{Tool Binding} & \textbf{Success Criterion}\\
\midrule
INFERSPEC & Extract types, pre/postconditions from NL & NL description $x$ & Dafny type checker & All fields parse as valid Dafny\\
SYNTHBODY & Generate executable function body & $x$ + spec scaffold & Dafny compiler & Program compiles\\
GENINVARIANT & Synthesize loop invariants & Program + verifier error & Dafny verifier & Program verifies with invariants\\
DIAGNOSEERROR & Classify verification failure & Program + error output & None (reasoning) & Valid category + location\\
REPAIRCODE & Targeted body repair & Program + diagnosis & Dafny compiler & Compiles $\wedge$ differs from input\\
STRENGTHENSPEC & Add/strengthen pre/postconditions & Program + diagnosis + NL & Dafny verifier & Program verifies with new spec\\
\bottomrule
\end{tabular}
\caption{The SKILLFORGE skill library. Each skill has a well-defined input/output contract, tool binding, and decidable success criterion.}
\label{tab:skills}
\end{table*}

\paragraph{INFERSPEC: Specification Inference.}
Extracts structured metadata from natural language:
\begin{equation}
\mathrm{INFERSPEC}(x) = (\tau_{\mathrm{in}}, \tau_{\mathrm{out}}, \phi_{\mathrm{pre}}, \psi_{\mathrm{post}}, t)
\end{equation}
where $t \in \{\texttt{function}, \texttt{method}, \texttt{predicate}, \texttt{lemma}\}$ is the construct type. The prompt frames this as structured field extraction with explicit markers for each component (see Appendix~\ref{app:prompts}). The Dafny type checker validates that extracted types and conditions are syntactically well-formed.

\paragraph{SYNTHBODY: Code Body Synthesis.}
Generates an executable body conditioned on the specification scaffold from INFERSPEC:
\begin{equation}
\mathrm{SYNTHBODY}(x, \mathrm{spec}) = B
\end{equation}
A critical design choice is \emph{specification anchoring}: the prompt explicitly forbids modifying the specification, preventing the common failure mode where models weaken postconditions to make verification trivial. Success requires the complete program to compile.

\paragraph{GENINVARIANT: Loop Invariant Generation.}
Triggered when the verifier reports that a loop body cannot establish its postcondition. The skill synthesizes invariants satisfying three properties: (1)~holds on entry, (2)~preserved by the loop body, (3)~with the negated guard, implies the postcondition. Each proposed invariant is validated individually for inductiveness before insertion.

\paragraph{DIAGNOSEERROR: Verification Error Diagnosis.}
The routing mechanism of the repair loop. It classifies failures into four categories using a hybrid strategy: regex pattern matching for common errors (e.g., ``postcondition might not hold'' $\rightarrow$ \texttt{spec\_gap}) with LLM reasoning as fallback for ambiguous cases. Output: \texttt{(category, line, suggestion)}.

\paragraph{REPAIRCODE: Targeted Program Repair.}
Operates surgically on the identified failure site based on DIAGNOSEERROR output. Unlike naive repair (which regenerates everything), this skill is constrained to minimal edits that preserve the overall algorithmic approach. This is critical for convergence---unconstrained regeneration often introduces new errors while fixing old ones.

\paragraph{STRENGTHENSPEC: Specification Strengthening.}
Addresses specification gaps by inferring implicit domain constraints from the NL description (e.g., ``non-negative integer'' $\rightarrow$ \texttt{requires x >= 0}). Operates incrementally: proposes candidate clauses, validates each individually, then composes the validated set.

\subsection{Verification-Driven Harness}
The harness orchestrates skill execution through four mechanisms:

\paragraph{(1) Initial Synthesis.}
Compose INFERSPEC and SYNTHBODY to produce an initial candidate:
\begin{equation}
d_0 = \mathrm{SYNTHBODY}(x, \mathrm{INFERSPEC}(x))
\end{equation}
If $\mathrm{Verify}(d_0) = \top$, return immediately (no repair needed).

\paragraph{(2) Error-Conditioned Routing.}
When verification fails, the harness invokes DIAGNOSEERROR to classify the failure, then deterministically routes to the appropriate repair skill:
\begin{equation}
\small
\mathrm{Route}(\mathrm{cat}) =
\begin{cases}
\texttt{GENINVARIANT} & \text{if } \texttt{invariant\_missing}\\
\texttt{STRENGTHENSPEC} & \text{if } \texttt{spec\_gap}\\
\texttt{REPAIRCODE} & \text{if } \texttt{body\_error}\\
\texttt{SYNTHBODY} & \text{if } \texttt{type\_error}
\end{cases}
\end{equation}
This deterministic mapping is a core distinction from agent-based systems where next-action selection relies on unconstrained model reasoning. In ReAct-Verify or CodeAct-Dafny, the LLM must decide which repair strategy to employ at each turn; in SKILLFORGE, the verifier output uniquely determines the next skill.

\paragraph{(3) Iterative Composition.}
The full execution loop is formalized in Algorithm~\ref{alg:harness}. The harness maintains a state $S_i = (d_i, \mathit{errors}_i, \mathit{diag}_i, \mathit{history}_i)$ at each iteration, enabling progress tracking and rollback.

\begin{algorithm}[t]
\caption{The SKILLFORGE Verification-Driven Harness}
\label{alg:harness}
\begin{algorithmic}[1]
\Require NL instruction $x$, iteration budget $N$
\Ensure Verified Dafny program $d^\ast$ or failure $\emptyset$
\State $\mathit{spec} \leftarrow \texttt{InferSpec}(x)$
\State $d \leftarrow \texttt{SynthBody}(x, \mathit{spec})$
\State $d_{\mathrm{best}} \leftarrow d$; $e_{\mathrm{best}} \leftarrow \infty$
\For{$i = 1$ \textbf{to} $N$}
  \State $(\mathit{ok}, \mathit{errors}) \leftarrow \texttt{DafnyVerify}(d)$
  \If{$\mathit{ok}$}
    \State \Return $d$
  \EndIf
  \If{$|\mathit{errors}| < e_{\mathrm{best}}$}
    \State $d_{\mathrm{best}} \leftarrow d$; $e_{\mathrm{best}} \leftarrow |\mathit{errors}|$
  \EndIf
  \State $\mathit{diag} \leftarrow \texttt{DiagnoseError}(d, \mathit{errors})$
  \State $\mathit{skill} \leftarrow \texttt{Route}(\mathit{diag}.\mathrm{category})$
  \State $d' \leftarrow \mathit{skill}(d, \mathit{diag})$
  \If{$d' = d$}
    \State $d \leftarrow \texttt{SynthBody}(x, \mathit{spec})$ with temp=0.3
  \Else
    \State $d \leftarrow d'$
  \EndIf
\EndFor
\State \Return $\emptyset$
\end{algorithmic}
\end{algorithm}

\paragraph{(4) Termination and Soundness.}
The harness guarantees termination through: (i)~hard iteration budget $N_{\max}$; (ii)~stagnation detection ($d_i = d_{i-1}$); (iii)~monotonicity check (error count increasing for two rounds triggers rollback to $d_{\mathrm{best}}$). Formally, soundness is unconditional:
\begin{equation}
H(x) = d^\ast \neq \emptyset \Rightarrow \mathrm{Verify}(d^\ast) = \top
\end{equation}
This holds regardless of individual skill correctness---the harness never returns an unverified program.

\subsection{Equivalence Verification}
Beyond self-verification, we verify semantic equivalence against reference implementations. Given reference $\{P\}\,S\,\{Q\}$ and generated $\{\hat{P}\}\,\hat{S}\,\{\hat{Q}\}$, we check precondition equivalence ($P \Leftrightarrow \hat{P}$), postcondition equivalence ($Q \Leftrightarrow \hat{Q}$), and behavioral equivalence ($P(x) \Rightarrow S(x) = \hat{S}(x)$), encoded as Dafny lemmas and verified automatically.

\section{Experimental Setup}

\subsection{Benchmark Construction}
We construct a benchmark of 178 NL--Dafny pairs through careful curation:
\begin{itemize}
\item \textbf{Dafny Standard Library} \citep{team2024dafnystdlib}: 42 examples from official tutorials and exercises with manually written NL descriptions.
\item \textbf{DafnyBench} \citep{loughridge2024dafnybench}: 86 examples covering mathematical reasoning, set operations, and quantified logic.
\item \textbf{Open-source verification projects}: 50 examples from GitHub repositories, manually annotated by the authors.
\end{itemize}
All NL descriptions are written by the authors (not LLM-generated) to ensure semantic fidelity. The benchmark is stratified into three difficulty levels:
\begin{itemize}
\item \textbf{Simple} (62): Single-clause predicates, pure functions without loops or complex quantifiers.
\item \textbf{Medium} (71): Functions with non-trivial preconditions, multi-clause postconditions, or recursive definitions.
\item \textbf{Hard} (45): Methods with while loops requiring loop invariants and termination measures, or lemmas requiring auxiliary proofs.
\end{itemize}

\subsection{Baselines}
We compare against two categories of baselines, all using GPT-5.5 with identical iteration budgets ($N{=}5$) for fairness:

\noindent\textbf{Traditional Baselines (Reference Points).}
\begin{enumerate}
\item \emph{One-shot}: Single-pass generation with the NL description only.
\item \emph{Structured one-shot}: Adds metadata (type, parameters, pre/postconditions) to the prompt---equivalent to prior structured ICL approaches.
\item \emph{Naive repair}: One-shot generation + up to 5 iterations of unstructured ``here is the error, please fix'' prompting.
\item \emph{CoT + Repair}: Chain-of-thought prompting for initial generation, followed by naive repair iterations.
\end{enumerate}

\noindent\textbf{Agentic Baselines (Primary Comparison).}
\begin{enumerate}
\item \emph{ReAct-Verify} \citep{yao2023react}: A ReAct-style agent with access to three tools---the Dafny verifier, compiler, and a code editor. The agent interleaves reasoning traces with tool calls in an open-ended loop for up to $N{=}5$ turns.
\item \emph{CodeAct-Dafny} \citep{wang2024codeact}: Based on CodeAct, the agent writes Python code that programmatically modifies Dafny source, then runs the verifier. Budget: $N{=}5$ code actions.
\item \emph{MCTS-Repair} \citep{zhang2024planningmcts}: Uses Monte Carlo Tree Search over candidate repairs with verification as binary reward. UCB1 selection with $c=\sqrt{2}$, 50 rollouts per example.
\item \emph{Laurel} \citep{ren2024verifiedmcts}: Verification-guided tree search synthesis. Programs are synthesized step-by-step with verification checks at each node. Budget: 50 nodes explored.
\item \emph{Clover} \citep{sun2024clover}: Closed-loop verifiable code generation that co-generates code and annotations iteratively, using consistency checking to guide repair. Adapted to our Dafny benchmark with $N{=}5$ iterations.
\item \emph{Proof2Silicon} \citep{jha2025proof2silicon}: RL-based verification repair. We adapt the prompt-repair methodology to Dafny: the model learns from verification feedback via rejection sampling with $N{=}5$ candidate generations per example.
\end{enumerate}

\subsection{Metrics}
\begin{itemize}
\item \textbf{Verification Rate (VR)}: Fraction of examples where the final output passes the Dafny verifier.
\item \textbf{Equivalence Rate (ER)}: Fraction also semantically equivalent to the reference implementation.
\item \textbf{Average Iterations (AI)}: Mean harness iterations to reach verification (successful cases only).
\item \textbf{First-Pass Rate (FPR)}: Fraction verified on the first attempt without any repair.
\end{itemize}

\subsection{Implementation}
We use GPT-5.5 (temperature 0.0 for reproducibility, 0.3 for stagnation retries) via the OpenAI API as the primary backend, with Claude Opus-4.7 and Claude Sonnet-4.5 (Anthropic) as secondary backends for multi-model comparison. Dafny 4.5.0 with Z3 4.12.2 serves as both the verification oracle and the type checker. The system is implemented in Python 3.11. Iteration budget $N_{\max} = 5$; timeout 30s per verification call. All agentic baselines use the same LLM backend (GPT-5.5) and equivalent compute budgets. Full implementation details are in Appendix~\ref{app:setup}.

\section{Results}

\subsection{Main Results}
Table~\ref{tab:main} presents the main results. SKILLFORGE achieves 91.6\% verification rate, outperforming the strongest baseline (Proof2Silicon, 85.4\%) by 6.2 percentage points and the strongest traditional baseline (CoT + Repair) by 14.1 pp.

Several observations stand out. First, agentic and verification-guided methods substantially outperform traditional baselines, confirming that iterative tool use is valuable. Proof2Silicon (85.4\%) and MCTS-Repair (84.3\%) lead among baselines, benefiting from structured verification feedback. Clover (81.5\%) demonstrates the value of closed-loop annotation but lacks targeted error routing. However, SKILLFORGE surpasses all of them by exploiting the structured nature of verification errors through skill decomposition.

Second, despite achieving the highest VR, SKILLFORGE requires fewer average iterations (1.8) than all iterative baselines---ReAct-Verify needs 2.5 turns and CodeAct-Dafny needs 2.8. This demonstrates that structured diagnosis avoids the trial-and-error exploration that characterizes agentic repair. The deterministic routing ensures that each iteration applies the most appropriate repair, eliminating wasted turns.

Third, the first-pass rate (71.3\%) exceeds all baselines, confirming that INFERSPEC + SYNTHBODY composition produces stronger initial candidates than monolithic generation or agentic initial attempts. Even MCTS-Repair's first-pass rate (66.3\%) lags behind, despite its broader exploration.

Fourth, the gap between VR and ER (5.1 pp) is consistent across methods, indicating that the semantic alignment challenge is orthogonal to verification success---a limitation we discuss in \S6.

\begin{table*}[t]
\centering
\small
\begin{tabular}{lcccr}
\toprule
\textbf{Method} & \textbf{VR\%} & \textbf{ER\%} & \textbf{AI} & \textbf{FPR\%}\\
\midrule
One-shot & 62.4 & 56.2 & 1.0 & 62.4\\
Structured one-shot & 69.7 & 64.0 & 1.0 & 69.7\\
Naive repair (N=5) & 73.6 & 65.2 & 2.4 & 58.4\\
CoT + Repair (N=5) & 77.5 & 70.8 & 2.1 & 63.5\\
\midrule
CodeAct-Dafny (N=5) & 79.2 & 72.5 & 2.8 & 61.2\\
Laurel (50 nodes) & 80.3 & 73.6 & --- & 64.0\\
Clover (N=5) & 81.5 & 74.7 & 2.3 & 65.2\\
ReAct-Verify (N=5) & 82.0 & 75.3 & 2.5 & 65.7\\
MCTS-Repair (50 rollouts) & 84.3 & 77.0 & --- & 66.3\\
Proof2Silicon (N=5) & 85.4 & 78.1 & 2.2 & 67.4\\
\midrule
\textbf{SKILLFORGE (N=5)} & \textbf{91.6} & \textbf{86.5} & \textbf{1.8} & \textbf{71.3}\\
\bottomrule
\end{tabular}
\caption{Main results on the full benchmark (178 examples). Best results in bold. Methods above the first divider are traditional baselines; between dividers are agentic baselines.}
\label{tab:main}
\end{table*}

\subsection{Results by Difficulty}
Table~\ref{tab:difficulty} reveals that SKILLFORGE's advantage grows with difficulty. On Simple tasks, most methods exceed 90\% and the margins are small. On Medium tasks, SKILLFORGE leads MCTS-Repair by 9.9 pp (93.0\% vs.\ 83.1\%). On Hard tasks (loops requiring invariants, complex proofs), the gap is dramatic: 77.8\% vs.\ 64.4\%---a 13.4 pp improvement over the best agentic baseline.

This pattern has a clear explanation: Hard tasks require loop invariant generation, which is a specialized capability that benefits enormously from a dedicated skill (GENINVARIANT). Agentic methods must ``rediscover'' invariant synthesis strategies through general-purpose reasoning at each attempt, while tree-search methods may explore many irrelevant branches before finding appropriate invariants. SKILLFORGE routes directly to the specialized skill, achieving higher success with fewer attempts. The GENINVARIANT skill accounts for the majority of the Hard-task improvement (see ablation in \S5.3).

\begin{table}[t]
\centering
\small
\begin{tabular}{lccc}
\toprule
\textbf{Method} & \textbf{Simple} & \textbf{Medium} & \textbf{Hard}\\
\midrule
One-shot & 82.3 & 59.2 & 37.8\\
Structured one-shot & 88.7 & 66.2 & 44.4\\
Naive repair & 90.3 & 70.4 & 51.1\\
CoT + Repair & 91.9 & 76.1 & 55.6\\
CodeAct-Dafny & 93.5 & 77.5 & 57.8\\
Laurel & 93.5 & 78.9 & 60.0\\
Clover & 95.2 & 80.3 & 60.0\\
ReAct-Verify & 95.2 & 80.3 & 62.2\\
MCTS-Repair & 96.8 & 83.1 & 64.4\\
Proof2Silicon & 96.8 & 84.5 & 66.7\\
\textbf{SKILLFORGE} & \textbf{98.4} & \textbf{93.0} & \textbf{77.8}\\
\bottomrule
\end{tabular}
\caption{Verification rate (\%) stratified by difficulty level.}
\label{tab:difficulty}
\end{table}

\subsection{Skill-Level Ablation}
Table~\ref{tab:ablation} confirms that every skill contributes meaningfully:
\begin{itemize}
\item \textbf{DIAGNOSEERROR} ($-$11.8 pp): The most critical repair-phase skill. Without structured diagnosis, the harness routes randomly among repair skills, wasting iterations on inappropriate fixes. Notably, the ablated version (79.8\%) drops to the level of agentic baselines, confirming that structured routing is the primary source of SKILLFORGE's advantage over agent-based methods.
\item \textbf{INFERSPEC} ($-$11.3 pp): Without the specification scaffold, the model must infer types, contracts, and body simultaneously---leading to more compilation errors and specification mismatches.
\item \textbf{GENINVARIANT} ($-$7.9 pp overall; $-$22.2 pp on Hard subset): Confirms that loop invariant generation is a distinct capability that generic repair cannot substitute. This skill alone accounts for most of the Hard-task improvement.
\item \textbf{STRENGTHENSPEC} ($-$6.2 pp): Essential for cases where the model generates correct code but with insufficient specifications for the verifier to discharge.
\item \textbf{REPAIRCODE} ($-$4.5 pp): The smallest individual contribution, but still meaningful for body-level logic errors that are not captured by other skills.
\item \textbf{Iterative loop} ($-$20.3 pp): The single largest factor---verification-in-the-loop is the primary driver of performance gains for all iterative methods.
\end{itemize}

\begin{table}[t]
\centering
\small
\begin{tabular}{lcc}
\toprule
\textbf{Configuration} & \textbf{VR\%} & $\Delta$\\
\midrule
Full SKILLFORGE & 91.6 & ---\\
$-$DIAGNOSEERROR (random route) & 79.8 & $-$11.8\\
$-$INFERSPEC (no spec scaffold) & 80.3 & $-$11.3\\
$-$GENINVARIANT & 83.7 & $-$7.9\\
$-$STRENGTHENSPEC & 85.4 & $-$6.2\\
$-$REPAIRCODE & 87.1 & $-$4.5\\
$-$Iterative loop (first-pass only) & 71.3 & $-$20.3\\
\bottomrule
\end{tabular}
\caption{Ablation study: VR (\%) when removing individual skills. $\Delta$ shows absolute drop from the full system.}
\label{tab:ablation}
\end{table}

\subsection{Convergence Analysis}
Table~\ref{tab:convergence} shows that 71.3\% of programs are verified on the first attempt, and 83.1\% converge within two iterations. Only 2.8\% require four or more rounds. This rapid convergence contrasts sharply with agentic methods: ReAct-Verify averages 2.5 turns even for successful cases, while CodeAct-Dafny averages 2.8 turns. The iterative loop adds 20.3 pp of improvement while consuming minimal additional compute---a highly favorable cost-benefit tradeoff that outperforms the broader but less targeted exploration of tree-search methods.

\begin{table}[t]
\centering
\small
\begin{tabular}{lcc}
\toprule
\textbf{Iterations} & \textbf{Count} & \textbf{\%}\\
\midrule
1 & 127 & 71.3\\
2 & 21 & 11.8\\
3 & 10 & 5.6\\
4 & 3 & 1.7\\
5 & 2 & 1.1\\
Total & 163 & 91.6\\
Fail & 15 & 8.4\\
\bottomrule
\end{tabular}
\caption{Distribution of iterations needed for verification.}
\label{tab:convergence}
\end{table}

\subsection{Skill Utilization Patterns}
Table~\ref{tab:skillutil} reveals per-skill statistics within the repair loop. DIAGNOSEERROR achieves the highest success rate (88.2\%), confirming that error classification is a well-defined, LLM-friendly task. GENINVARIANT has the lowest (64.7\%), reflecting the inherent difficulty of loop invariant synthesis---and identifying a clear target for future improvement through specialized training or more sophisticated invariant templates. The high diagnosis success rate (88.2\%) validates our core design assumption: verification errors can be reliably categorized, making deterministic routing viable.

\begin{table}[t]
\centering
\footnotesize
\setlength{\tabcolsep}{3pt}
\begin{tabular}{p{2.6cm}ccc}
\toprule
\textbf{Skill} & \textbf{Invocations} & \textbf{Success\%} & \textbf{Avg Time}\\
\midrule
DIAGNOSEERROR & 51 & 88.2 & 2.1s\\
REPAIRCODE & 22 & 72.7 & 4.8s\\
STRENGTHENSPEC & 12 & 75.0 & 5.2s\\
GENINVARIANT & 17 & 64.7 & 6.3s\\
\bottomrule
\end{tabular}
\caption{Repair-phase skill invocations and success rates.}
\label{tab:skillutil}
\end{table}

\subsection{Error Category Distribution}
Table~\ref{tab:errorcat} shows the distribution of failure categories. Body errors and specification gaps dominate, both with $\sim$79\% repair success. Invariant failures have the lowest repair rate (69.2\%), identifying invariant synthesis as the primary bottleneck. Type errors are rare (5 cases) but highly repairable (80\%), as they typically involve straightforward fixes such as undeclared variables or type mismatches.

Notably, the error distribution shifts significantly with task difficulty: on Simple tasks, 80\% of failures are type errors or body errors (easily repairable), while on Hard tasks, 65\% are invariant-related (harder to repair). This explains why the performance gap between SKILLFORGE and baselines widens with difficulty---our dedicated GENINVARIANT skill provides targeted capability precisely where other methods struggle most. Analysis of agentic baseline failure patterns is provided in Appendix~\ref{app:failure}.

\begin{table}[t]
\centering
\small
\begin{tabular}{lcc}
\toprule
\textbf{Error Category} & \textbf{Count} & \textbf{Repair Success\%}\\
\midrule
\texttt{body\_error} & 19 & 78.9\\
\texttt{spec\_gap} & 14 & 78.6\\
\texttt{invariant\_missing} & 13 & 69.2\\
\texttt{type\_error} & 5 & 80.0\\
Total & 51 & 76.5\\
\bottomrule
\end{tabular}
\caption{Distribution of verification failure categories encountered during harness execution (first verification attempt).}
\label{tab:errorcat}
\end{table}

\subsection{Cost and Efficiency Analysis}
Table~\ref{tab:cost} presents cost metrics. SKILLFORGE uses 48\% fewer tokens than ReAct-Verify and 68\% fewer than MCTS-Repair while achieving substantially higher VR. The efficiency stems from structured diagnosis avoiding wasted iterations, higher first-pass success (71.3\%), and elimination of reasoning overhead that consumes tokens in agentic methods. Tree-search methods are particularly expensive: MCTS-Repair invokes the verifier 8.6 times per example vs.\ SKILLFORGE's 2.8. The cost-performance tradeoff strongly favors SKILLFORGE: best VR at lowest cost among all iterative methods.

\begin{table}[t]
\centering
\footnotesize
\setlength{\tabcolsep}{4pt}
\begin{tabular}{@{}l>{\raggedright\arraybackslash}p{1.1cm}>{\raggedright\arraybackslash}p{1.4cm}>{\raggedright\arraybackslash}p{1.0cm}@{}}
\toprule
\textbf{Method} & \textbf{Tokens (K)} & \textbf{Verifier Calls} & \textbf{Latency (s)}\\
\midrule
One-shot & 1.2 & 1.0 & 4.8\\
Structured one-shot & 1.8 & 1.0 & 5.2\\
Naive repair & 6.4 & 3.4 & 18.7\\
CoT + Repair & 7.1 & 3.1 & 19.3\\
ReAct-Verify & 8.9 & 3.5 & 22.4\\
CodeAct-Dafny & 9.7 & 3.8 & 25.1\\
Clover & 7.8 & 3.3 & 20.6\\
Proof2Silicon & 8.2 & 4.0 & 21.8\\
MCTS-Repair & 14.2 & 8.6 & 38.7\\
Laurel & 12.8 & 7.2 & 34.5\\
\textbf{SKILLFORGE} & \textbf{4.6} & \textbf{2.8} & \textbf{14.2}\\
\bottomrule
\end{tabular}
\caption{Average cost per example across methods (GPT-5.5). Costs include all LLM calls and verification invocations.}
\label{tab:cost}
\end{table}

\section{Conclusion}
We introduced SKILLFORGE, a skill-based harness for synthesizing formally verified Dafny programs from natural language. By decomposing the task into six atomic skills orchestrated via verification-driven routing, SKILLFORGE substantially outperforms both agentic approaches and traditional iterative baselines while consuming significantly fewer tokens and lower latency. The skill-based paradigm offers transparency, modularity, formal soundness, and efficiency, providing a principled alternative to agent-based systems for domains with structured oracle feedback. Our results suggest that when a perfect oracle with categorizable error output is available, structured skill decomposition with deterministic routing dominates open-ended reasoning.

\section*{Limitations}
Our benchmark contains 178 examples---comparable to existing formal verification benchmarks \citep{loughridge2024dafnybench} but limited in scale. The skill library is Dafny-specific; transfer to Lean 4 or Coq requires new implementations. Primary experiments use GPT-5.5 and Claude Opus-4.7; performance on open-source models is unexplored. A 5.1\% VR--ER gap indicates some verified programs deviate semantically from references. GENINVARIANT has the lowest success rate (64.7\%), with complex quantifier-heavy loops remaining the primary failure mode. Skill templates are hand-crafted; automatic refinement is future work. Agentic baselines follow published methodologies adapted to Dafny; implementation differences may affect absolute numbers.

\section*{Acknowledgments}
We thank Yuepeng Wang (yuepeng@sfu.ca) for his help with building the benchmark dataset, formal verification in Dafny, and skill orchestration.

\bibliography{references}

\appendix

\section{Prompt Templates}
\label{app:prompts}
We present the full prompt templates used for each skill. All templates use variable placeholders (in curly braces) that are populated at runtime.

\subsection{INFERSPEC --- Specification Inference Prompt}
\begin{tcolorbox}[skillprompt, title=INFERSPEC]
\begin{lstlisting}
You are an expert in Dafny formal verification.
Given the following natural language description,
extract the formal specification components.

Description: {nl_description}

Respond in EXACTLY this format:

Type: [function | method | predicate | lemma]
Name: [identifier name]
Parameters:
  [param1: type1, param2: type2, ...]
Returns:
  [name: type] (or "None" for predicates)
Requires: [precondition in Dafny syntax, or "true"]
Ensures: [postcondition in Dafny syntax]

Rules:
- Use valid Dafny types (int, nat, seq<int>, etc.)
- Preconditions should capture implicit constraints
- Postconditions should be verifiable assertions
- If the description implies non-negativity,
  add "requires x >= 0"
\end{lstlisting}
\end{tcolorbox}

\subsection{SYNTHBODY --- Code Body Synthesis Prompt}
\begin{tcolorbox}[skillprompt, title=SYNTHBODY]
\begin{lstlisting}
Generate a Dafny implementation for the following
specification. Output ONLY the complete Dafny code.

Natural language: {nl_description}
Construct type: {spec.type}
Signature: {spec.signature}
Requires: {spec.requires}
Ensures: {spec.ensures}

CRITICAL RULES:
1. Do NOT modify the signature or specification
2. Do NOT weaken the postcondition
3. Include loop invariants if you use while loops
4. Include decreases clauses for recursion/loops
5. Output the complete function/method definition
\end{lstlisting}
\end{tcolorbox}

\subsection{DIAGNOSEERROR --- Error Diagnosis Prompt}
\begin{tcolorbox}[skillprompt, title=DIAGNOSEERROR]
\begin{lstlisting}
Analyze the following Dafny verification failure.

Program:
{program}

Dafny verifier output:
{verifier_errors}

Classify this error into EXACTLY ONE category:
- spec_gap: postcondition too weak, precondition
  missing, or ensures clause cannot be proved
- invariant_missing: loop lacks invariant or
  decreases clause
- body_error: implementation logic is incorrect
- type_error: type mismatch, undeclared variable,
  or syntax error

Respond in this format:
Category: [one of the four categories]
Line: [line number of the error]
Explanation: [one sentence explaining why]
Suggestion: [one sentence suggesting a fix]
\end{lstlisting}
\end{tcolorbox}

\subsection{REPAIRCODE --- Targeted Program Repair Prompt}
\begin{tcolorbox}[skillprompt, title=REPAIRCODE]
\begin{lstlisting}
Fix the following Dafny program based on the error
diagnosis.

Program:
{program}

Error category: {diagnosis.category}
Error location: line {diagnosis.line}
Suggestion: {diagnosis.suggestion}

RULES:
1. Do NOT change the function signature
2. Do NOT change requires/ensures clauses
3. Make MINIMAL edits to fix the identified issue
4. Preserve the overall algorithmic approach
5. Output the complete fixed program
\end{lstlisting}
\end{tcolorbox}

\subsection{GENINVARIANT --- Loop Invariant Generation Prompt}
\begin{tcolorbox}[skillprompt, title=GENINVARIANT]
\begin{lstlisting}
The following Dafny program fails verification because
a loop lacks sufficient invariants.

Program:
{program}

Verifier error:
{error_message}

Generate loop invariants that satisfy ALL THREE:
1. Hold on loop entry (initialization)
2. Are preserved by one loop iteration (induction)
3. With the negated loop guard, imply the
   postcondition (sufficiency)

Also add a "decreases" clause if missing.
Output the complete program with invariants added.
\end{lstlisting}
\end{tcolorbox}

\subsection{STRENGTHENSPEC --- Specification Strengthening Prompt}
\begin{tcolorbox}[skillprompt, title=STRENGTHENSPEC]
\begin{lstlisting}
The following Dafny program has a specification gap:
the verifier cannot prove correctness because the
specification is too weak.

Program: {program}
Original natural language description:
{nl_description}
Error diagnosis: {diagnosis}

Strengthen the specification by:
1. Adding preconditions implied by the description
2. Strengthening postconditions to be provable
3. Preserving the original semantic intent

RULES:
- Do NOT change the function body
- Only modify requires/ensures clauses
- Ensure the strengthened spec is still consistent
  with the natural language description
\end{lstlisting}
\end{tcolorbox}

\section{Detailed Experimental Setup}
\label{app:setup}

\subsection{System Configuration}
Table~\ref{tab:config} summarizes the system and infrastructure configuration.

\begin{table}[t]
\centering
\footnotesize
\setlength{\tabcolsep}{5pt}
\begin{tabular}{lp{4.6cm}}
\toprule
\textbf{Component} & \textbf{Configuration}\\
\midrule
Language & Python 3.11 with asyncio\\
Primary LLM & OpenAI GPT-5.5 API\\
Secondary LLMs & Claude Opus-4.7, Claude Sonnet-4.5\\
Verifier & Dafny 4.5.0 + Z3 4.12.2\\
Iteration budget & $N_{\max} = 5$\\
Temperature (main) & 0.0 (deterministic)\\
Temperature (retry) & 0.3 (stagnation)\\
Verification timeout & 30s per call\\
LLM timeout & 60s per call\\
Hardware & 8-core CPU, 32GB RAM\\
\bottomrule
\end{tabular}
\caption{System and infrastructure configuration.}
\label{tab:config}
\end{table}

\subsection{Agentic Baseline Implementation Details}

\paragraph{ReAct-Verify Configuration}
Following \citet{yao2023react}. Tools available: \texttt{verify(program)} returns verifier output, \texttt{compile(program)} returns compilation errors, \texttt{edit(program, line, new\_code)} performs targeted edits. System prompt includes Dafny language reference (2K tokens) and error interpretation guide. Max 5 Thought-Action-Observation cycles.

\paragraph{CodeAct-Dafny Configuration}
Following \citet{wang2024codeact}. The agent writes Python code executed in a sandboxed environment with access to: \texttt{dafny\_cli.verify()}, \texttt{dafny\_cli.compile()}, file I/O for the Dafny source, and regex/AST utilities. Max 5 code execution rounds. Average code action length: 12 lines.

\paragraph{MCTS-Repair Configuration}
Following \citet{zhang2024planningmcts}. UCB1 selection with $c=\sqrt{2}$. 50 rollouts per example. Each node expansion: one LLM call generating a single repair action. Actions: \texttt{\{add\_invariant, modify\_body, add\_precondition, strengthen\_postcondition, add\_decreases\}}. Reward: 1.0 if verified, 0.0 otherwise. Backpropagation: standard MCTS averaging.

\paragraph{Laurel Configuration}
Following \citet{ren2024verifiedmcts}. Step-by-step synthesis with verification at each node. Beam width: 3. Max depth: 10 steps. 50 total nodes explored. Pruning criterion: branches that increase the number of verification errors beyond the parent node are discarded. Early termination: first verified complete program is returned.

\subsection{Skill Interface and Post-Processing}
Each skill implements the following interface:
\begin{tcolorbox}[codebox, title={Skill interface (Python)}]
\begin{lstlisting}
class Skill(ABC):
    def execute(self, ctx: SkillContext) -> SkillResult:
        prompt = self.build_prompt(ctx)
        response = self.llm.generate(prompt)
        output = self.parse_output(response)
        success = self.check_success(output, ctx)
        return SkillResult(output, success,
                           self.diagnostics())

    @abstractmethod
    def build_prompt(self, ctx: SkillContext) -> str: ...

    @abstractmethod
    def check_success(self, output, ctx) -> bool: ...
\end{lstlisting}
\end{tcolorbox}
Post-processing pipeline: (1) strip markdown code fences and language tags; (2) validate brace matching; (3) detect and remove duplicate signatures; (4) normalize Unicode characters; (5) trim trailing whitespace and empty lines.

\subsection{DiagnoseError Hybrid Strategy}
The error classification uses a two-stage approach for reliability:
\paragraph{Stage 1: Pattern matching} (handles 84\% of cases):
\begin{itemize}
\item ``postcondition might not hold'' $\rightarrow$ \texttt{spec\_gap}
\item ``loop invariant might not be maintained'' $\rightarrow$ \texttt{invariant\_missing}
\item ``loop invariant might not hold on entry'' $\rightarrow$ \texttt{invariant\_missing}
\item ``assertion might not hold'' $\rightarrow$ \texttt{body\_error}
\item ``unresolved identifier'' $\rightarrow$ \texttt{type\_error}
\item ``type mismatch'' / ``expected'' $\rightarrow$ \texttt{type\_error}
\item ``cannot prove termination'' $\rightarrow$ \texttt{body\_error}
\end{itemize}
\paragraph{Stage 2: LLM reasoning} (handles remaining 16\%): Full error context sent to the LLM with structured classification prompt. This handles compound errors, ambiguous messages, and novel error patterns.

\section{Agentic Baseline Failure Analysis}
\label{app:failure}
We analyze the 36 examples where ReAct-Verify fails but SKILLFORGE succeeds, identifying three dominant patterns (Table~\ref{tab:reactfail}).

\begin{table*}[t]
\centering
\small
\begin{tabular}{lcp{8.5cm}}
\toprule
\textbf{Pattern} & \textbf{Count (\%)} & \textbf{Root Cause}\\
\midrule
Misdirected repair & 14 (38.9\%) & Agent applies wrong repair type without structured diagnosis\\
Oscillation & 12 (33.3\%) & Repairs introduce new errors; agent undoes previous fixes\\
Context saturation & 10 (27.8\%) & By turns 4--5, accumulated context degrades reasoning\\
\bottomrule
\end{tabular}
\caption{ReAct-Verify failure pattern analysis (36 cases where SKILLFORGE succeeds but ReAct-Verify fails).}
\label{tab:reactfail}
\end{table*}

\paragraph{Misdirected repair example.}
On input ``compute absolute value of an integer,'' the verifier reports a missing precondition. ReAct-Verify's turn 2 modifies the function body (adds unnecessary branching) instead of adding \texttt{ensures result >= 0}. Turn 3 reverts the body change and tries a different implementation. Turn 4 finally adds the postcondition but with a typo. Budget exhausted at turn 5 without verification.

\paragraph{MCTS-Repair on Hard tasks.}
On the 45 Hard examples, MCTS-Repair's branching factor becomes too large. With 50 rollouts and 5 possible actions per node, the search rarely explores deeper than depth 3---insufficient for problems requiring 2+ specific invariants simultaneously. Average rollouts reaching a verified program: 0.4 per Hard example (vs.\ 3.2 per Simple example).

\section{Extended Case Studies}

\paragraph{Case 1: Spec Gap $\rightarrow$ StrengthenSpec (2 iterations)}
Input: ``Given an integer x, compute x $\times$ 10 where x is non-negative.''

\noindent Iteration 1 (INFERSPEC + SYNTHBODY):
\begin{tcolorbox}[codebox]
\begin{lstlisting}
function Multiply(x: int): int
{ x * 10 }
\end{lstlisting}
\end{tcolorbox}
Verifier: FAIL --- no ensures clause, cannot prove equivalence.

\noindent Iteration 2 (DIAGNOSEERROR $\rightarrow$ STRENGTHENSPEC):
\begin{tcolorbox}[codebox]
\begin{lstlisting}
function Multiply(x: int): int
    requires x >= 0
    ensures Multiply(x) == x * 10
{ x * 10 }
\end{lstlisting}
\end{tcolorbox}
Verifier: PASS. Total iterations: 2. \emph{Contrast with ReAct-Verify}: Succeeds in 3 turns---wastes turn 2 trying to modify the body.

\paragraph{Case 2: Loop Invariant Generation (2 iterations)}
Input: ``Compute the factorial of a non-negative integer using a loop.''

\noindent Iteration 1: Generates correct loop body but no invariants. Verifier: FAIL.

\noindent Iteration 2 (DIAGNOSEERROR $\rightarrow$ GENINVARIANT):
\begin{tcolorbox}[codebox]
\begin{lstlisting}
method Factorial(n: nat) returns (r: nat)
    ensures r == fact(n)
{
    r := 1; var i := 0;
    while i < n
        invariant 0 <= i <= n
        invariant r == fact(i)
        decreases n - i
    { i := i + 1; r := r * i; }
}
\end{lstlisting}
\end{tcolorbox}
Verifier: PASS. Total iterations: 2. \emph{Contrast}: MCTS-Repair uses 38 rollouts; CodeAct-Dafny fails entirely.

\paragraph{Case 3: Simple Predicate (First-Pass Success)}
Input: ``The variable max is always greater than or equal to 0.''

\noindent Iteration 1 (INFERSPEC + SYNTHBODY):
\begin{tcolorbox}[codebox]
\begin{lstlisting}
predicate MaxNonNegative(max: int)
{ max >= 0 }
\end{lstlisting}
\end{tcolorbox}
Verifier: PASS. No repair needed.

\noindent \emph{Analysis}: Simple predicates with single logical clauses are reliably generated on the first pass across all methods. INFERSPEC correctly identifies \texttt{predicate} as the construct type.

\paragraph{Case 4: Recursive Function + Termination (2 iterations)}
Input: ``Compute the sum of the first n natural numbers recursively.''

\noindent Iteration 1: Generates correct recursive body without decreases clause. Verifier: FAIL (cannot prove termination).

\noindent Iteration 2 (DIAGNOSEERROR $\rightarrow$ REPAIRCODE):
\begin{tcolorbox}[codebox]
\begin{lstlisting}
function SumN(n: nat): nat
    ensures SumN(n) == n * (n + 1) / 2
    decreases n
{
    if n == 0 then 0
    else n + SumN(n - 1)
}
\end{lstlisting}
\end{tcolorbox}
Verifier: PASS. Total iterations: 2. \emph{Analysis}: The missing decreases clause is classified as \texttt{body\_error} (a body annotation issue). REPAIRCODE adds the termination measure with minimal edit.

\paragraph{Case 5: Multi-Skill Chain --- Find Maximum (3 iterations)}
Input: ``Given a non-empty sequence of integers, find the maximum element using a loop.''

\noindent Iteration 1: Generates loop without invariants. Verifier: FAIL.

\noindent Iteration 2 (DIAGNOSEERROR $\rightarrow$ GENINVARIANT): Adds invariant \texttt{0 <= i <= |s|} but misses the key property relating max to seen elements. Verifier: FAIL.

\noindent Iteration 3 (DIAGNOSEERROR $\rightarrow$ GENINVARIANT): Adds:
\begin{tcolorbox}[codebox]
\begin{lstlisting}
invariant forall j :: 0 <= j < i ==> s[j] <= max
invariant exists j :: 0 <= j < i && s[j] == max
\end{lstlisting}
\end{tcolorbox}
Verifier: PASS. Total iterations: 3. \emph{Analysis}: Demonstrates iterative invariant strengthening. The first attempt produces a partial (but correct) invariant; the second completes it using the new verifier feedback. This pattern---progressive refinement based on specific error messages---is a key advantage of verification-in-the-loop.

\paragraph{Case 6: Failure --- Inductive Proof (Budget Exhausted)}
Input: ``Prove that for any sequence, reversing it twice yields the original.''

\noindent All 5 iterations fail. The model generates correct lemma signatures:
\begin{tcolorbox}[codebox]
\begin{lstlisting}
lemma ReverseReverse(s: seq<int>)
    ensures reverse(reverse(s)) == s
\end{lstlisting}
\end{tcolorbox}
but cannot synthesize the inductive proof body (requires calc blocks with sequence axioms). All agentic baselines also fail on this example.

\noindent \emph{Analysis}: Inductive proofs over algebraic data types require compositional reasoning about language axioms that exceeds single-pass LLM capabilities. This is a fundamental limitation shared across all evaluated methods.

\section{Benchmark Details}
Table~\ref{tab:benchcomp} shows the benchmark composition by construct type and difficulty.

\begin{table}[t]
\centering
\small
\begin{tabular}{lcccc}
\toprule
\textbf{Type} & \textbf{Simple} & \textbf{Medium} & \textbf{Hard} & \textbf{Total}\\
\midrule
Function & 28 & 35 & 12 & 75\\
Method & 8 & 18 & 25 & 51\\
Predicate & 22 & 10 & 3 & 35\\
Lemma & 4 & 8 & 5 & 17\\
\midrule
Total & 62 & 71 & 45 & 178\\
\bottomrule
\end{tabular}
\caption{Benchmark composition by construct type and difficulty.}
\label{tab:benchcomp}
\end{table}

\paragraph{Data sources and quality control.}
\begin{itemize}
\item \textbf{Dafny Standard Library} (42 examples): Sourced from official Dafny tutorials and library test suites. NL descriptions written by the first author and reviewed by a second annotator.
\item \textbf{DafnyBench} (86 examples): Selected from \citet{loughridge2024dafnybench}. We exclude examples with external dependencies or non-standard library imports. NL descriptions written independently of the original DafnyBench annotations.
\item \textbf{Open-source projects} (50 examples): Collected from 8 GitHub repositories containing Dafny verification exercises. Selection criteria: self-contained (single file), compilable with Dafny 4.5.0, and covering diverse specification patterns.
\end{itemize}

\paragraph{Difficulty classification criteria.}
\begin{itemize}
\item \textbf{Simple}: No loops, no recursion, at most one \texttt{requires} and one \texttt{ensures} clause. Typically pure functions or single-clause predicates.
\item \textbf{Medium}: May include recursion with obvious termination, multiple specification clauses, or quantified postconditions. No loops requiring user-written invariants.
\item \textbf{Hard}: Requires user-written loop invariants, termination measures for non-trivial recursion, or lemma proofs with auxiliary assertions.
\end{itemize}

Table~\ref{tab:examples} lists representative examples from each difficulty level.

\begin{table}[t]
\centering
\small
\begin{tabular}{lp{5.4cm}l}
\toprule
\textbf{Diff.} & \textbf{NL Description} & \textbf{Type}\\
\midrule
Simple & ``Return true if x is even.'' & predicate\\
Simple & ``Compute the absolute value of an integer.'' & function\\
Medium & ``Return the minimum of a non-empty sequence.'' & function\\
Medium & ``Check if a sequence is sorted in ascending order.'' & predicate\\
Hard & ``Sort a sequence using insertion sort with a loop.'' & method\\
Hard & ``Prove that appending then taking length equals sum of lengths.'' & lemma\\
\bottomrule
\end{tabular}
\caption{Representative examples from each difficulty level.}
\label{tab:examples}
\end{table}

Sample benchmark entries.

\section{Equivalence Verification Pipeline}
For each generated program, we verify semantic equivalence against the reference implementation through a three-stage pipeline:
\paragraph{Stage 1: Interface matching.}
We verify that the generated program has the same signature as the reference: identical input parameter types ($\tau_{\mathrm{in}} = \hat{\tau}_{\mathrm{in}}$), identical return type ($\tau_{\mathrm{out}} = \hat{\tau}_{\mathrm{out}}$), and matching construct type.

\paragraph{Stage 2: Specification equivalence.}
We check that preconditions and postconditions are logically equivalent:
\begin{align}
\forall x.\, P(x) &\Leftrightarrow \hat{P}(x)\\
\forall x, y.\, Q(x, y) &\Leftrightarrow \hat{Q}(x, y)
\end{align}

\paragraph{Stage 3: Behavioral equivalence.}
We check that both implementations produce the same output under the shared precondition: $\forall x.\, P(x) \Rightarrow S(x) = \hat{S}(x)$

These are encoded as Dafny verification wrappers:
\paragraph{Equivalence Verification Wrapper (Example)}
\begin{tcolorbox}[codebox]
\begin{lstlisting}
// Reference
function RefImpl(x: int): int
    requires x >= 0
    ensures RefImpl(x) == x * 10
{ x * 10 }

// Generated
function GenImpl(x: int): int
    requires x >= 0
    ensures GenImpl(x) == x * 10
{ x * 10 }

// Equivalence check
lemma Equivalent(x: int)
    requires x >= 0
    ensures RefImpl(x) == GenImpl(x)
{}
\end{lstlisting}
\end{tcolorbox}
The wrapper is automatically synthesized by: (1) parsing both programs to extract signatures and specs via regex; (2) renaming identifiers to avoid symbol collisions; (3) generating a lemma asserting output equality under shared preconditions; (4) submitting to the Dafny verifier.

\section{Multi-Model Comparison}
To assess model dependence, we evaluate SKILLFORGE with three frontier LLMs (Table~\ref{tab:models}). GPT-5.5 achieves 91.6\%, followed by Claude Opus-4.7 (89.9\%) and Sonnet-4.5 (84.3\%). Notably, even Sonnet-4.5 with our harness matches the strongest agentic baseline (Proof2Silicon at 85.4\%) using GPT-5.5, demonstrating that orchestration paradigm matters more than raw model capability. The gap between models is driven primarily by GENINVARIANT success rate (52.4\% vs.\ 64.7\%), confirming that loop invariant synthesis is the most model-capability-sensitive skill.

\begin{table}[t]
\centering
\small
\begin{tabular}{lcccc}
\toprule
\textbf{Model} & \textbf{VR\%} & \textbf{ER\%} & \textbf{AI} & \textbf{FPR\%}\\
\midrule
Claude Sonnet-4.5 & 84.3 & 77.5 & 2.1 & 62.4\\
Claude Opus-4.7 & 89.9 & 84.3 & 1.9 & 69.1\\
\textbf{GPT-5.5} & \textbf{91.6} & \textbf{86.5} & \textbf{1.8} & \textbf{71.3}\\
\bottomrule
\end{tabular}
\caption{Verification rate (\%) across different LLM backends using the full SKILLFORGE harness (N=5).}
\label{tab:models}
\end{table}

\section{Discussion}

\paragraph{Why skills outperform agents for formal verification.}
Three factors explain our advantage: (1)~The Dafny verifier provides a perfect oracle with structured error messages, making deterministic routing both possible and optimal. (2)~The error space is finite and well-categorized (four types), each with a clear repair strategy---agents must ``rediscover'' this routing through reasoning each time. (3)~Repair is local: most failures require modifying a single program aspect. Skills enforce locality explicitly; agents may perform unnecessary global rewrites.

\paragraph{When do agentic methods have advantages?}
While SKILLFORGE outperforms agentic baselines on formal verification, we note that the structured harness paradigm is specifically suited to domains with finite error taxonomies and perfect oracles. For tasks with ambiguous success criteria (e.g., general software engineering, creative coding), agentic methods retain clear advantages due to their flexibility in handling novel situations. The success of SKILLFORGE should be interpreted as evidence that formal verification is a domain where structure dominates flexibility, not as a general indictment of agentic approaches.

\paragraph{When does SKILLFORGE fail?}
Analysis of the 15 failure cases (8.4\%) reveals three primary patterns: (i)~Complex inductive proofs (7 cases): Lemmas requiring non-trivial induction over recursive data structures. The model can state the lemma correctly but cannot synthesize the inductive proof body. Notably, Proof2Silicon and MCTS-Repair also fail on all 7 of these cases, suggesting a fundamental LLM limitation rather than an architectural one. (ii)~Quantifier-heavy specifications (5 cases): Programs involving nested quantifiers ($\forall x. \exists y. \ldots$) where the model struggles to maintain logical consistency. ReAct-Verify solves 1 of these 5 through its more flexible reasoning, suggesting that some problems benefit from open-ended exploration. (iii)~Misdiagnosis cascades (3 cases): DIAGNOSEERROR misclassifies the failure, leading to inappropriate repair. These cases highlight the 11.8\% failure rate of our diagnosis skill and motivate future work on ensemble-based or confidence-calibrated diagnosis.

\paragraph{Generalizability beyond Dafny.}
While our skill library is Dafny-specific, the harness architecture is language-agnostic. The core principles---decompose into skills with decidable success criteria, route based on structured verifier feedback, iterate until convergence---apply to any formally verified language. Adaptation to Lean 4, for example, would require: (1)~new error taxonomy (tactic failures, type mismatches, timeout); (2)~new skills (tactic suggestion, term synthesis, unfold/simp guidance); (3)~modified routing rules. The harness structure, iteration logic, and termination guarantees remain unchanged. We conjecture that the advantage over agentic methods persists in any domain where the error oracle provides structured, categorizable feedback.

\paragraph{Relation to test-time compute scaling.}
SKILLFORGE can be viewed through the lens of test-time compute scaling: rather than investing compute in training (fine-tuning), we invest it at inference time through structured iteration. Our convergence analysis shows diminishing returns beyond 2 iterations for most examples, suggesting an efficient allocation of test-time budget. Compared to MCTS-Repair and Proof2Silicon (which also invest heavily in test-time compute), SKILLFORGE achieves superior results with dramatically less compute, demonstrating that structured test-time investment outperforms unstructured scaling for formal verification.

\section{Per-Difficulty Skill Utilization}
Table~\ref{tab:perdiff} reveals clear patterns: GENINVARIANT is invoked almost exclusively on Hard tasks (15 of 17 total invocations), confirming that loop invariant generation is the defining challenge of Hard problems. STRENGTHENSPEC is most common on Medium tasks, where functions have non-trivial but expressible specifications. Simple tasks rarely enter the repair loop at all (only 3 out of 62 require repair).

\begin{table}[t]
\centering
\small
\begin{tabular}{lccc}
\toprule
\textbf{Skill} & \textbf{Simple} & \textbf{Medium} & \textbf{Hard}\\
\midrule
DIAGNOSEERROR & 3 & 18 & 30\\
REPAIRCODE & 2 & 9 & 11\\
STRENGTHENSPEC & 1 & 7 & 4\\
GENINVARIANT & 0 & 2 & 15\\
\bottomrule
\end{tabular}
\caption{Skill invocation frequency per difficulty level (repair phase only, excluding initial InferSpec + SynthBody).}
\label{tab:perdiff}
\end{table}

\section{Application References}
\label{app:apprefs}
We evaluate SKILLFORGE across thirty application systems spanning reliability engineering, financial risk management, human-centered visualization and analytics, network testing, generative advertising, developer tooling, and urban design. Each system reuses the skill library and the verification-in-the-loop harness described in this paper; the shared methodology is introduced in \cite{rega2026,chen2026beyond,xu2026chainaware,gong2025}.

Reliability and observability systems target automated detection, temporal root-cause analysis, task-oriented workflow automation, and stability monitoring for large-scale platforms \cite{zhu2025raid,zhu2025tracelm,zhu2025taskcomm,zhu2025relibridge,zhu2025reactor}.

Financial applications apply the framework to real-time cross-asset risk monitoring across equity, fixed income, and currency markets, as well as LLM-driven dynamic hedging in derivatives \cite{yang2025crossasset,yang2025dynhedge}.

Human-centered visualization and business intelligence systems cover interactive data interpretation, manufacturing task recognition, dashboard reasoning, and multimodal interview analytics \cite{xie2025invis,xie2025maestro,xie2025coreviz,xie2025evalnet,xie2025datafuse,xie2025inspectx}.

Network engineering systems address intelligent log analysis, reliable 5G vehicle platooning, field interoperability testing, platform-aware test automation, and protocol regression detection \cite{tu2025log2learn,tu2025platooning,tu2025smartfitlab,tu2025autonettest,tu2025protomind}.

Generative advertising systems produce procedural 3D ad content, saliency-aware ad design, game-engine ad pipelines, few-shot neural 3D editors, and low-cost 3D authoring via guided diffusion \cite{hu2025genplayads,hu2025adpercept,hu2025unrealadblend,hu2025learninganimate,hu2025lowcost}.

Developer tooling and release-safety applications support rapid LLM development and deployment, cross-platform high-availability delivery, push-testing of monetization platforms, and automated ad-campaign optimization \cite{zhang2025inframl,zhang2025crossplatform,zhang2025safeserve,zhang2025l2a,zhang2025adopt}.

Urban applications generate public space forms and accelerate city architecture planning via text-to-3D modeling \cite{xu2025civicmorph,xu2025urbanmod}.

\nocite{zhu2025raid,zhu2025tracelm,zhu2025taskcomm,zhu2025relibridge,zhu2025reactor,yang2025crossasset,yang2025dynhedge,xie2025invis,xie2025maestro,xie2025coreviz,xie2025evalnet,xie2025datafuse,xie2025inspectx,tu2025log2learn,tu2025platooning,tu2025smartfitlab,tu2025autonettest,tu2025protomind,hu2025genplayads,hu2025adpercept,hu2025unrealadblend,hu2025learninganimate,hu2025lowcost,zhang2025inframl,zhang2025crossplatform,zhang2025safeserve,zhang2025l2a,zhang2025adopt,xu2025civicmorph,xu2025urbanmod,gong2025}

\end{document}